\pdfoutput=1

\documentclass[11pt]{article}

\usepackage[preprint]{acl}

\usepackage{times}
\usepackage{latexsym}

\usepackage{subfigure}

\usepackage[T1]{fontenc}

\usepackage[utf8]{inputenc}

\usepackage{microtype}

\usepackage{inconsolata}

\usepackage{graphicx}
\usepackage{url}
\usepackage{listings}
\usepackage{stfloats}  

\title{Text Classification in the LLM Era - Where do we stand?}

\author{Sowmya Vajjala \\
  National Research Council \\
  Ottawa, Canada \\
    \texttt{sowmya.vajjala@nrc-cnrc.gc.ca} \\\And
  Shwetali Shimangaud \\
  McGill University \\
  Montreal, Canada \\
  \texttt{shwetali.shimangaud@mail.mcgill.ca } \\}

\begin{document}
\maketitle
\begin{abstract}
Large Language Models revolutionized NLP and showed dramatic performance improvements across several tasks. In this paper, we investigated the role of such language models in text classification and how they compare with other approaches relying on smaller pre-trained language models. Considering 32 datasets spanning 8 languages, we compared zero-shot classification, few-shot fine-tuning and synthetic data based classifiers with classifiers built using the complete human labeled dataset. Our results show that zero-shot approaches do well for sentiment classification, but are outperformed by other approaches for the rest of the tasks, and synthetic data sourced from multiple LLMs can build better classifiers than zero-shot open LLMs. We also see wide performance disparities across languages in all the classification scenarios. We expect that these findings would guide practitioners working on developing text classification systems across languages. 
\end{abstract} 

\section{Introduction}
Text classification is one of the evergreen problem in NLP and other related areas of research, with widespread applications across different real-world use cases and disciplines of study. Each classification use case is different, and collecting sufficient labeled data for each problem can be challenging. This resulted in the interest in the development of zero-shot text classification systems \cite{yin2019benchmarking}. Large Language Models can offer a solution, and their use as zero-shot (English) text classifiers \citep{gretz2023zero} has been explored in recent past. Synthetic data generation with LLMs has also been proposed to address the labeled data scarcity. Different from these two approaches, there is an established body of work on few-shot fine-tuning \citep[e.g.,][]{tunstall2022efficient,yehudai2024fastfit}, to address situations where we have a very small amount of labeled data, which is not typically sufficient to build a classifier using standard methods. 

How does zero-shot text classification compare with few-shot fine-tuning, synthetic data based classification, and building classifiers with full labeled datasetshough some parts of this broad question received attention in the recent past, a more comprehensive comparison is lacking in NLP research. Further, most of the past research in this direction has focused on English datasets and proprietary LLMs. A detailed comparison across different classification methods spanning more languages and datasets will not only help us understand the state of the art in text classification with LLMs, but also provide guidance to practitioners looking at solving real-world text classification use cases across various tasks and languages. We address these issues in this paper.  

Concretely, we study the following research questions in this paper, considering 32 datasets covering 8 languages (4 datasets per language). 
\begin{enumerate}
\item How well does zero-shot prompting of LLMs (open and proprietary) fare compared to building classifiers with full training data?
\item Does few-shot fine-tuning offer any benefits over zero-shot classification? 
\item How well does a synthetic data based classifier fare compared to zero-shot classification with LLMs?
\item Is supervised/instruction fine-tuning of LLMs the way to go for text classification? 
\end{enumerate}


Starting with an overview of related work (Section ~\ref{sec:related}), we proceed to the description of our methodology (Section ~\ref{sec:approach}) and discuss the details about the results (Section ~\ref{sec:results}). After summarizing the main conclusions (Section ~\ref{sec:disc}), we discuss the limitations (Section ~\ref{sec:limitations}) and broader impacts (Section ~\ref{sec:broad}). 

\section{Related Work}
\label{sec:related}
Text classification, the task of classifying a given text into a pre-defined list of categories, is a well-studied problem. From bag-of-words features to the current state-of-the-art LLMs, numerous approaches have been explored in the past. Access to large amounts of human labeled data has traditionally played a significant role in improving text classifiers, and NLP research in the past two decades addressed this issue by looking at different solutions to learn from little or no labeled data.

\paragraph{Zero-shot Pre-LLM Approaches: }Some of the earlier classification approaches relied on using only label names to build "data less" text classifier \cite{liu2004text,li2018pseudo,meng2020text,ye2020zero,gera-etal-2022-zero} and embedding the texts and labels in a shared space \cite{song2014dataless,luo2021don,chu2021unsupervised,sarkar-etal-2022-exploring,gao-etal-2023-benefits,wang-etal-2023-pesco}. \citet{yin2019benchmarking} proposed to formulate zero-shot text classification as a textual entailment problem, although \citet{ma2021issues} point to the limitations of this approach in terms of variability across datasets and reliance on spurious lexical patterns. Another practical approach for zero-shot classification is cross-lingual transfer i.e., train a classification model in one or more languages, and use it as a zero-shot classifier on the target language \cite{wang-banko-2021-practical}. Except \citep{wang-banko-2021-practical}, who studied sentiment and hate speech classification tasks, all the research has  focused primarily on English datasets. 

\paragraph{Few-shot fine-tuning: } Approaches that can learn from a small amount of ($<20$ samples per category) labeled examples have also been explored in the recent years \cite{schick2020automatically,dopierre-etal-2021-neural,ohashi2021distinct,zhang-etal-2022-prompt-based}. SetFit \cite{tunstall2022efficient} introduced an approach based on supervised contrastive learning, transforming a language model into a topic encoder using only a few examples per label, and demonstrated effectiveness with datasets where the number of categories are low (under 5). FastFit \cite{yehudai2024fastfit} proposed an approach that scales to many classes (50--150) effectively, and showed its usefulness with English datasets. Out of these only SetFit evaluated with a few non-English datasets. 

\paragraph{Zero-shot Classification with LLMs: }With the arrival of Large Language Models, some recent approaches explored proprietary models like GPT3.5 and GPT4 for zero-shot or few-shot in-context learning for text classification across several datasets \cite{ gretz2023zero,sun-etal-2023-text,mozes-etal-2023-towards,tian-chen-2024-esg}. Extending this line of work, open LLMs were studied in the context of intent classification \cite{ruan-etal-2024-large,arora-etal-2024-intent} and computational social science \cite{mu2024navigating}. However, comparing such zero-shot approaches with few-shot and full-data  based fine-tuning, \cite{edwards-camacho-collados-2024-language} show that smaller, fine-tuned classifiers outperform zero-shot approaches. Whether supervised fine-tuning of LLMs offers any benefit is an unexplored question. Surprisingly, except \citep{tian-chen-2024-esg}, all these experiments have been focused only on English datasets so far. We expand this strand of work to 7 other languages, and provide more detailed comparisons across different LLMs. 

\paragraph{Synthetic Data: } One approach to address the labeled data problem is to augment existing data by creating new data by applying text transformations such as replacing synonyms, paraphrasing, back translation etc. \citet{bayer2022survey} presents a detailed survey of such data augmentation techniques for text classification. An extension of this idea is to directly synthesize the labeled data using generative language models \cite{yu-etal-2023-regen,yue-etal-2023-synthetic,kurakin2023harnessing, choi-etal-2024-unigen}. In the recent past, Large Language Model based synthetic data generation is increasingly observed across different NLP tasks \cite{tan-etal-2024-large}. GPT4 has been used for English \cite{li2023synthetic,yamagishi-nakamura-2024-utrad,peng-etal-2024-incubating} and code-mixed \cite{zeng-2024-leveraging} synthetic data generation for text classification with mixed results. We extend this line of work by covering more languages and exploring multiple LLMs as sources for synthetic data instead of relying on one, and extending to handle datasets with a larger label set. 

Overall, we address several gaps in existing research by comparing zero-shot classification, few-shot fine-tuning, synthetic data based classification, and classification with full data together, and also study how the comparison works out once we go beyond English. In this process, we also present a comparison between different open and closed recent LLMs. 

\section{Approach}
\label{sec:approach}

We experimented with zero-shot classification, few-shot fine-tuning, and synthetic data based classification, and compared them with classifiers trained on full amount of labeled data. Our methods are described below, followed by a description of the datasets used. 

\subsection{Zero-shot Prompting}
We compared three open LLMs - Qwen2.5-7B \cite{qwen2.5}, Aya23-8B \cite{aryabumi2024aya} and Aya-Expanse-8B \cite{dang2024aya}, which is a more recent, instruction tuned version of Aya23, and one proprietary LLM - GPT4 \cite{achiam2023gpt} (\textit{gpt-4-0613}) in a zero-shot prompting setup across all languages and classification tasks. Initial experiments showed a tendency to generate a lot of explanation for the prediction despite specifying not to in the prompt. So, we controlled for the output structure using Instructor.\footnote{\url{https://python.useinstructor.com/}} Further details on Instructor setup are mentioned in the Appendix (Figure \ref{fig:prompt}). All LLMs still generated explanations beyond labeling, (as high as 10\% for some open LLMs) which were treated as classification errors. All prompts were in English, as changing the language to the target language of the dataset resulted in poorer results in early experiments, which was also observed in some recent studies on other problems/datasets \cite{dey2024better,jin2024better}. We did not attempt few-shot prompting, considering the large label set with some of the datasets, but looked into few-shot fine-tuning, instead, as described below. 

\subsection{Few-shot fine-tuning} 
We performed few-shot fine-tuning with FastFit \cite{yehudai2024fastfit} which integrates batch contrastive learning with a token similarity score to learn few-shot task specific representations for text classification. We used 10 examples per label in all cases, as that had the best result in the original FastFit paper.\footnote{Base model: \href{https://huggingface.co/sentence-transformers/paraphrase-multilingual-mpnet-base-v2}{paraphrase-multilingual-mpnet-base-v2}} We experimented with another few-shot fine-tuning approach SetFit \cite{tunstall2022efficient} but it quickly became intractable to train for some of the datasets with $>$10 categories. Hence, we reported results with only FastFit in this paper. Comparisons with SetFit for the datasets with under $10$ categories can be seen in the Appendix (Section~\ref{app:moreresults}). 

\subsection{Synthetic Data Generation}
We generated equal amounts of synthetic data from three sources - GPT4, Qwen2.5-7B and Aya-Expanse-8B, for all the classification tasks, across all languages, to ensure diversity in the generated text. Initial experiments showed that generating data from multiple LLMs was beneficial than relying on a single source, which is corroborated by recent research on other tasks \cite{maheshwari2024efficacy}. This is also useful for controlling the costs, as the two open LLMs can be run locally on a laptop and do not incur any inference costs (and consumed less power). We used the same prompt across all LLMs, changing the task/language as needed. Details about the prompting strategy can be seen in Appendix  ~\ref{app:settings}. 

\subsection{Classification with Synthetic and Real Data}
We compared three approaches for text classification with real or synthetic training data, listed below:
\begin{enumerate}
\item A logistic regression classifier with the embedding representations from a state-of-the-art transformer model as the feature vector generator, without any further fine-tuning. We used \textit{gte-multilingual-base} \cite{zhang-etal-2024-mgte}, a 305 million parameter multilingual model, as our feature extractor. 
\item A fine-tuned version of BERT \cite{devlin-etal-2019-bert} with multilingual BERT as the base,\footnote{\url{https://huggingface.co/google-bert/bert-base-multilingual-cased}} trained for 5 epochs, across all languages and datasets.  
\item Instruction fine-tuning of Qwen-2.5-7B-Instruct \cite{qwen2,qwen2.5} for 3 epochs on the training data (10 epochs for \textsc{Taxi1500}, the smallest dataset) for all languages.\footnote{Early experiments showed superior performance with Qwen compared to Aya-Expanse}\end{enumerate}
All the classifiers were trained in two setups: first only with real data, and then only with synthetic data. More details on the experimental setup such as parameters, time taken to train, GPU requirements etc are described in the Appendix (Section~\ref{app:settings}).

\subsection{Datasets and Evaluation}
We experimented with four publicly available datasets and each dataset has eight language subsets for Arabic, English, French, German, Hindi, Italian, Portuguese and Spanish (i.e, 32 datasets in total) with official train-validation-test splits. Arabic and Hindi datasets are in their native scripts and all the other languages are in Roman script. Our choice of datasets primarily depended on finding all languages represented across all datasets. The datasets cover sentiment and topic classification, and are described below:

\begin{enumerate}
\item Multilingual twitter sentiment \citep{barbieri2022xlm} which we will refer to as \textsc{sentiment} is a dataset of tweets manually labeled with positive/negative/neutral sentiment.
\item Taxi 1500 \citep{ma2023taxi1500} is a topic classification dataset, manually labelled with 6 categories - recommendation, faith, description, sin, grace and violence that describe sentences from the bible. The dataset covers 1500 languages in total, with a mapping between parallel sentences across bible versions that is used to build a labeled dataset from English labeled data. Some languages have multiple bibles, and we took the alphabetically first bible for that language to build our dataset. 
\item Amazon Massive \citep{fitzgerald-etal-2023-massive} is a one million sample dataset covering 51 languages consisting of parallel virtual assistant commands classified into 60 intents spread across 18 domains ("scenario" field in the dataset). We modeled intent and scenario classification as two separate tasks, which we refer to as \textsc{Intent} and \textsc{Scenario} datasets respectively.
\end{enumerate}

Note that all the datasets contain short texts of different genres (tweets, bible sentences and commands to voice assistants). Table~\ref{tab:datasetstats} shows a summary of the datasets used. 

\begin{table}[htb!]
    \centering
    \begin{tabular}{|l|l|l|l|} \hline
    Dataset & \# categories & \# train & \# test \\ \hline
      \textsc{sentiment}   & 3 & 1839 & 870\\
       \textsc{Taxi1500}  & 6 & 860 & 111 \\
        \textsc{Scenario} & 18 & 12000 & 2974 \\
       \textsc{Intent} & 60 & 12000 & 2974\\ \hline
    \end{tabular}
    \caption{Dataset statistics per language}
    \label{tab:datasetstats}
\end{table}

For synthetic data generation, we aimed to generate datasets comparable to the size of the training data for all dataset-language combinations except in the case of \textsc{Taxi1500} where we generated a training set that is double the size of the original human labeled training set owing to its small size compared to others. Table~\ref{tab:syndatasets} shows the sizes of the generated datasets and the split between different LLMs (GPT4, Aya-Expanse-8B, Qwen2.5-7B). 

\begin{table}[htb!]
    \centering
    \begin{tabular}{|l|l|l|} \hline
    Dataset & \# train & description \\ \hline
      \textsc{sentiment}   & 1800 & 200 per category, per LLM\\
       \textsc{Taxi1500}  & 1800 & 100 per category, per LLM \\
       \textsc{Intent} & 13500 & 75 per category, per LLM \\
       \textsc{Scenario} & 13500 & from intent dataset \\ \hline
    \end{tabular}
    \caption{Synthetic training data (per language)}
    \label{tab:syndatasets}
\end{table} 

\paragraph{Evaluation: } We report classification accuracy as the evaluation measure in this paper. Since two of the datasets are imbalanced across categories (\textsc{Taxi1500} and \textsc{Scenario}), we considered reporting macro-F1 additionally. But considering the fact that there is not much difference between the measures and the order is always preserved (i.e., if approach A gets a higher accuracy than approach B, it always has a higher macro-F1 as well), we decided to report only accuracy. 

\section{Results}
\label{sec:results}
We report results addressing the four research questions and also discuss the variation across languages and tasks in this section. Detailed per language/per task/per method results can be seen in Appendix ~\ref{app:moreresults}. 

\subsection{Zero-shot Classification}
\begin{figure}[htb!]
  \includegraphics[width=\columnwidth]{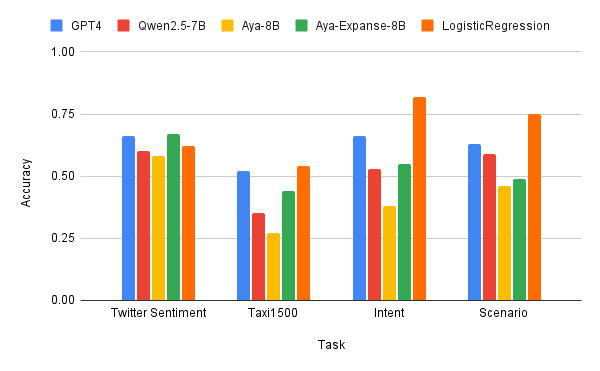}
  \caption{Zero-shot LLMs versus a logistic regression classifier trained with full data}
  \label{fig:zeroshot}
\end{figure}

Figure~\ref{fig:zeroshot} shows the zero-shot performance of various LLMs, compared to a logistic regression classifier trained with full data and embeddings based feature representation, averaged across all languages per task. 

These results reveal some interesting insights: For sentiment classification, GPT4 and Aya-Expanse are better than a classifier trained on full training data. But, for the other three tasks, GPT4 is clearly much better among the zero-shot methods, although we observe that a custom classifier is much better, especially as the number of labels in the dataset increases. The difference in performance trends between sentiment classification versus other tasks we studied here may indicate a more subjective versus topical task difference which would warrant further scrutiny. Interestingly, all the models performed poorly on French sentiment classification compared to other languages, while Arabic was the language where most models performed the worst for other tasks (see Tables ~\ref{tab:twitter-details}--~\ref{tab:scenario-details} in Appendix~\ref{app:moreresults} for details). 

\subsection{Few-shot Classification}
\begin{figure}[htb!]
  \includegraphics[width=\columnwidth]{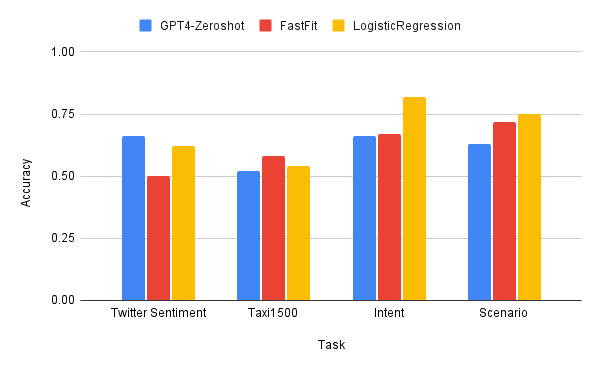}
  \caption{Few-shot Fine-tuning}
  \label{fig:fewshot}
\end{figure}

Figure~\ref{fig:fewshot} shows how few-shot fine-tuning with FastFit compares with zero-shot classification and training with full data, taking GPT4 as the representative zero-shot classifier, as that was the best among the zero-shot options we explored.

The results show that while few-shot fine-tuning is not useful for sentiment classification, there is a >5\% improvement over zeroshot GPT4 for two tasks (\textsc{Taxi1500} and \textsc{Scenario}) and similar performance for \textsc{Intent} datasets. For \textsc{Taxi1500}, it even results in a small improvement over training with full labeled dataset, presumably due to the contrastive learning objective used for learning the representations for fine-tuning. While there are performance disparities across languages (See Figure~\ref{fig:fewshotdetails} in Appendix~\ref{app:moreresults} for details), they are much larger for the datasets with a smaller number of categories (\textsc{Sentiment} and \textsc{Taxi1500}) compared to the datasets with larger number of categories (\textsc{Scenario} and \textsc{Intent}).Some of this can be attributed to the fact that the datasets with more categories see a larger sample of data during fine-tuning (as we take 10 samples per category), which is probably helping the model learn the task better across languages, reducing disparities among them in terms of overall accuracy. While we would need further experiments with other datasets with many classes (covering multiple languages), we can conclude based on these results that few-shot fine-tuning can be a viable alternative to zero-shot classifiers if at least a small amount of labeled data is available. 

\subsection{Synthetic Data and Text Classification}
\label{subsec:zerovssyn}

\begin{figure}[htb!]
  \includegraphics[width=\columnwidth]{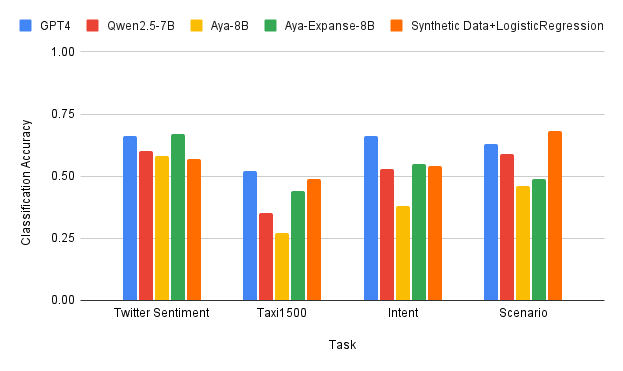}
  \caption{Zero-shot versus synthetic data based Classification}
  \label{fig:zerovssyn1}
\end{figure}

We now turn to the question of the usefulness of synthetic data for text classification. Figure~\ref{fig:zerovssyn1} shows a comparison between all the zero-shot LLMs and a logistic regression model trained entirely with synthetic data, averaged across languages and grouped by task.

In all tasks except sentiment classification, we notice that the synthetic data based classifier either performs comparably or out-performs all the open LLMs and outperforms GPT4 too in one task (\textsc{Scenario}). We can infer that synthetic data can be considered a viable option over zero-shot classification, from these results. In practical terms, that can mean a one time cost (for building the synthetic dataset) rather than an ongoing cost of prompting a proprietary LLM as a zero-shot classifier instead. 

\begin{figure}[htb!]
  \includegraphics[width=\columnwidth]{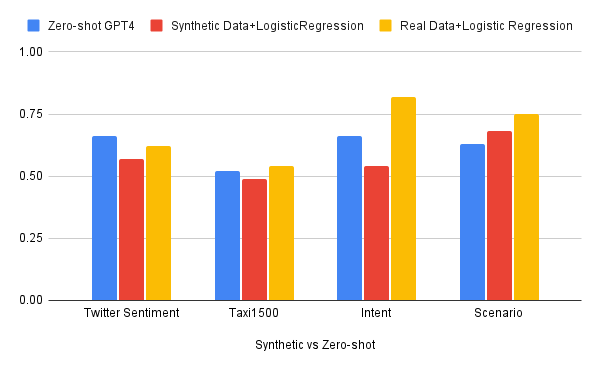}
  \caption{zero-shot GPT4, synthetic data based, and real data based classifiers}
  \label{fig:zerovssyn2}
\end{figure}

Figure~\ref{fig:zerovssyn2} compares among zero-shot, synthetic data based, and real data based classifiers, taking GPT4 as the zero-shot classifier. 

We can observe from this figure that for at least one task \textsc{Taxi1500}, synthetic data is performing at the same level as the other two approaches, whereas for the \textsc{Scenario} task, it show slightly better results than zero-shot GPT4. Considering Figures ~\ref{fig:zerovssyn1} and ~\ref{fig:zerovssyn2} together, we can conclude that synthetic data generated from multiple LLMs can be useful and sometimes, better than zero-shot classifiers, while being competitive compared to real-data in some cases. 

\paragraph{One versus Many Synthetic Data Sources}
Since we used three sources for synthetic data generation, a natural next question to look into is what is a good source of data. To understand this, we compared different sources of synthetic data by  using only one source to build a logistic regression based classifier each time. Table~\ref{tab:synbysource} shows the summary of these results. Note that these single-source datasets form only 1/3rd of the full dataset which uses all three sources together. Hence, we don't compare with classifiers trained on full data here, and compare only one LLM versus another as a source of synthetic data. 
\begin{table}[htb!]
    \centering 
    \begin{tabular}{|p{1.5cm}|p{0.9cm}|p{2.1cm}|p{1.3cm}|} \hline
    Task  & GPT4 & Aya-Expanse & Qwen2.5 \\ \hline 
    \textsc{Twitter} & 0.5 & 0.47 & \textbf{0.51} \\
    \textsc{Taxi1500} & \textbf{0.43} & 0.4 & \textbf{0.43} \\ 
    \textsc{Intent}&\textbf{0.53}&0.48&0.40\\ 
    \textsc{Scenario} & \textbf{0.65}& 0.64& 0.59\\ \hline
    \end{tabular}
    \caption{Average accuracy across languages of synthetic data based classification}
    \label{tab:synbysource}
\end{table} 

We can see that Qwen2.5 gave better results for the two datasets with smaller number of categories, but started to perform the worse of the three LLMs once we moved to datasets with larger number of categories. GPT4 consistently seems to be a reasonable source of synthetic data across all datasets. Aya-Expanse does better with datasets with larger number of categories than smaller ones. Training on the data from all sources consistently gives better results despite these differences in individual sources (Figure~\ref{fig:zerovssyn1}). Thus, we can conclude that multi-source generation also potentially results in more diverse data and using open LLMs as the sources of synthetic data along with GPT4 can be a cost effective way of synthetic data generation. Note that the open LLMs are both very small (7B/8B) compared to GPT4 and can be used locally on a laptop. 

\begin{figure*}[htb!]
  \centering
  \subfigure[real data]{\includegraphics[width=0.45\textwidth]{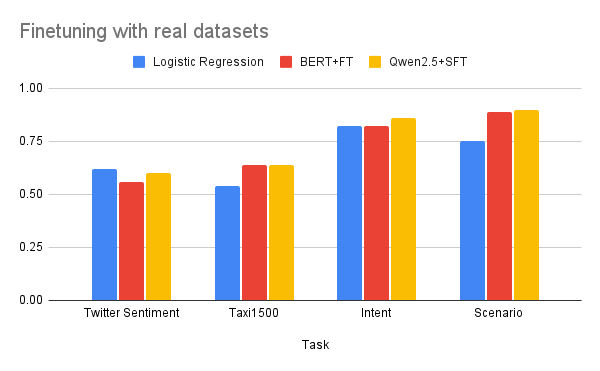}}
  \subfigure[synthetic data]{\includegraphics[width=0.45\textwidth]{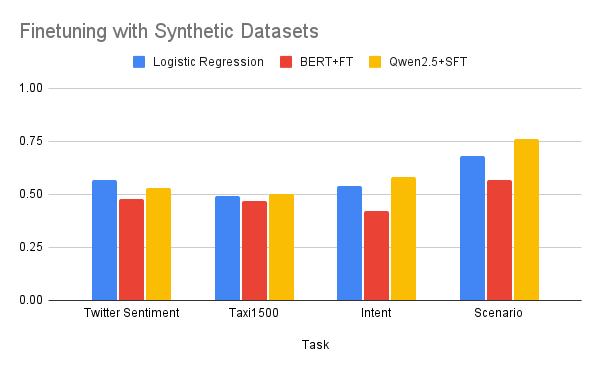}}
  \caption{Synthetic versus Real-Data}
  \label{fig:synvsreal}
\end{figure*}

\paragraph{Classification with Real versus Synthetic Data}
We compared three classification approaches: a logistic regression classifier, which we used in all the above described experiments to compare against zero-shot and few-shot approaches, a classifier fine-tuned on the multilingual BERT model, and an instruction tuned classifier built from the Qwen2.5-7B-Instruct model. Figure \ref{fig:synvsreal} presents the performance of these classifiers using both real and synthetic datasets. With real datasets, we can see the plain logistic regression model give the best average performance for sentiment classification over all languages. It falls behind other approaches (although not dramatically) with other tasks. The Qwen2.5 and BERT fine-tuned models achieved similar accuracy across most tasks, except for the \textsc{Intent} dataset, where Qwen finetuned model outperformed BERT by 5\%. On the synthetic datasets, BERT fine-tuned model consistently had lower accuracy across tasks. The Qwen finetuned model showed the best performance for datasets with a large number of labels (\textsc{Intent} and \textsc{Scenario}). In summary, our results indicate that instruction tuning is perhaps more effective for synthetic datasets and tasks with more categories, whereas logistic regression remains a strong choice for simpler tasks with fewer categories.  

\subsection{Performance Differences Across Languages}
Across all tasks and methods, we noticed large differences in performance across languages in these experiments. Table~\ref{tab:perfdiff} shows the performance difference between the best and worst performing languages for all methods averaged across the four tasks. Exact language specific and task specific details can be seen in the Appendix (Section~\ref{app:moreresults}). 

\begin{table}[htb!]
    \centering
    \begin{tabular}{|l|l|} \hline
    Method & Difference\\ \hline
      Zero-shot GPT4 & 13\%\\
      Zero-shot Qwen2.5-7B & 15\%\\
      Zero-shot Aya-8B & 13\%\\
      Zero-shot Aya-Expanse-8B & 11\%\\ 
      FastFit & 22\%\\
      Logistic Regression-Real & 29\%\\
    BERT-FT-Real & 23\%\\
    SFT-Real & 17\%\\
      Logistic Regression-Synthetic & 17\%\\
    BERT-FT-Synthetic & 12\%\\
    SFT-Synthetic & 17\%\\ \hline
    \end{tabular}
    \caption{Average (Absolute) Performance Difference Across Languages for the different methods}
    \label{tab:perfdiff}
\end{table} 

Zero-shot prompting, followed by synthetic data based classification had relatively less variation across languages, although they are still in the 10-20\% range. When we had some labeled data for all languages, we still had $>20\%$ performance difference for few-shot finetuning and BERT finetuning, whereas a logistic regression classifier has an almost 30\% average difference between best and worst performing languages. In specific cases, such as few-shot finetuning on \textsc{Taxi1500}, the difference between the best performing (French-69\%) and least performing (Arabic-30\%) was nearly 40\%. Even when there was full access to training data, BERT fine-tuning had $>40\%$ performance difference between Italian (35\%) and English (77\%) for \textsc{Taxi1500}. 

Although this dataset has the least amount of training data, we observe similar trend in datasets where large amounts of training data is available. For example, the logistic regression classifier on \textsc{Scenario} dataset has 58\% accuracy with Arabic, but 90\% with English. All these may point to the insufficiency of the base multilingual embedding models in representing data across languages. It is also possible that the difference in the script and the writing direction is a contributing factor in some cases (e.g., Arabic). 

While the variance across languages is lesser with synthetic data, as we noticed earlier, multilingual synthetic data generation comes with other challenges. For example, generation is much slower for languages with a different script (Arabic and Hindi in our data). The average number of tokens was also higher (in some cases, 3 times higher) for non-Roman script based languages. This directly influences the costs involved in synthetic data generation, especially with proprietary LLMs, and may impose limitations on their use as synthetic data generators for problems involving data from non roman script based languages. 

\section{Discussion}
\label{sec:disc}
We compared different ways of performing text classification (zero-shot classification, few-shot fine-tuning, using full labeled data, and using synthetic data) across 32 datasets (8 languages, 4 datasets per language). Returning to our original research questions, our findings are summarized below:

\paragraph{RQ1: } \textit{How well does zero-shot prompting of LLMs (open and proprietary) fare compared to building classifiers with full training data?}

Zero-shot classifiers perform well in terms of accuracy, but primarily for datasets with fewer categories, especially \textsc{sentiment}, where GPT4 and an open LLM Aya-Expanse-8B achieve comparable results which are better than training with full human labeled data. In all other cases, while GPT4 has the best performance among the zero-shot LLMs, it trails behind classifiers built with labeled datasets, especially as the number of categories increases. With high amounts of labeled data, even a logistic regression classifier with text embedding representations performs much better, and is of course less resource and cost intensive. 


\paragraph{RQ2: } \textit{Does few-shot fine-tuning offer any benefits over zero-shot classification?}

Few-shot fine-tuning generally offers higher accuracy (upto 10\% gain) than zero-shot classifiers on 3 out of 4 tasks. However, for the \textsc{sentiment} task, which has fewer categories and can be perceived as a more subjective compared to topic classification, GPT-4 outperforms few-shot finetuning. Overall, considering the fast training and inference times, without any added costs, few-shot finetuning can be considered a reliable option as the number of categories increases, in the absence of sufficient labeled data.  

\paragraph{RQ3: } \textit{How well does a synthetic data based classifier fare compared to zero-shot classification with LLMs?}

In all tasks except sentiment classification, the synthetic data-based classifier either performs on par (\textsc{intent}) or achieves a 5-10\% improvement compared to zero-shot classification using open LLMs. It outperforms GPT-4 too in one task (\textsc{scenario}). These results indicate that synthetic data can be a viable alternative to zero-shot classification with open LLMs when number of categories are more, as it involves a one-time cost for creating the dataset. In terms of what is a better source of synthetic data, in most cases, GPT4 and at least one open LLM achieve comparable performance as the single source of synthetic data. Over all, the best results are achieved by combining all data sources, which can also be a cost-effective solution.   


\paragraph{RQ4: } \textit{Is supervised fine-tuning of LLMs the way to go for text classification?}

Excluding sentiment classification, we see that the best performance is with either BERT fine-tuned or an instruction fine-tuned classifier with real data, and supervised fine-tuning does better than other approaches with synthetic data. However, it has to be noted that supervised fine-tuning is compute intensive both for training and inference across all languages and tasks (See Table~\ref{tab:parameter-setting-for-FT} in the appendix for details on the time taken). On the other hand, we also notice a strong performance with a simple logistic regression based classifier too in some cases, although as the number of categories increases, the advantage seems to wane away.


Based on these results, we can summarize the following guidelines for practitioners:
\begin{enumerate}
\item For sentiment classification, zero-shot classification with LLMs is a better option than task specific fine-tuning.
\item In all other cases, few-shot fine-tuning achieves better performance than zero-shot classification. So, collecting a handful of labeled data is useful.
\item Synthetic data based classifiers perform better than zero-shot classification with open LLMs, but are not always better than GPT4. However, sourcing data from multiple LLMs is useful. 
\item Despite all the recent advances, training classifiers using high-quality labeled data still gives the best performance, and SFT gives the best performance, especially when dealing with a large number of categories, and even a logistic regression classifier can give a strong performance in some cases when such datasets are available. 
\end{enumerate}

While the datasets and languages covered are by no means exhaustive, these results provide some guidance on what methods work for what kinds of data, what to expect in terms of language disparities, and how to work with synthetic data from multiple sources, across multiple languages. Future directions can include increasing the coverage of languages, and expanding to multi-label datasets, to draw more comprehensive conclusions about LLMs and the task of text classification. 

\section{Limitations}
\label{sec:limitations}
While the study allows us to draw a few concrete conclusions based on our experiments, it is not free from limitations. Firstly, we have limited ourselves to one prompt for zero-shot models and one prompt for synthetic data generation. While the use of instructor for structured output generation in the case of zero-shot classification automatically adjusts the prompt during retries, essentially taking care of prompt engineering, we cannot steer the internal prompt creation process ourselves. Few-shot prompting (instead of few-shot fine-tuning) was also not explored, as the benefits are unclear with the increasing number of labels in some datasets. Few-shot fine-tuning can be sample sensitive, and while we did not notice any variations across different runs, we did not systematically explore that aspect. We also did not explore multilingual classifiers and all our classifiers are monolingual, built on top of pre-trained multilingual models. No language-specific choices were made (e.g., using a English embedding model may give better performance for English). We also did not do any qualitative analysis of the results or of the generated synthetic data. 

Since our goal is to compare broadly across different approaches, an extensive evaluation of fine-tuning options or parameter variations was not performed, nor did we repeat the experiments with different initializations, to keep the number of experiments (and costs involved) in check. Additionally, all the datasets deal with only short texts (tweets, sentences, voice assistant commands) and hence, the results of this study might not extend to long texts. Finally, the question of potential data contamination is inevitable while discussing the zero-shot performance. While we don't know the specifics of the training data for various LLMs, considering that the performance across all tasks (except sentiment classification) is lower than models relying on full training data, perhaps it is not a serious concern for these experiments. In terms of the languages covered, while there is some typological diversity, 7/8 languages belong to the Indo-European language family (covering three sub-families: Germanic, Italic and Indo-Aryan). Thus, the extendability of these conclusions to other languages and language families is not guaranteed. Finally, the open LLMs we explored are much smaller in size (7B-8B parameters) compared to GPT4, and the results should not be seen as a verdict against the use of open LLMs for text classification. 

\section{Ethics and Broader Impact}
\label{sec:broad}
We have used publicly available datasets and did not do any experiments involving human participants. We have also used small, locally run LLMs for several experiments, and most of the experiments are run locally on a laptop (details in the appendix), thus, consuming less power and perhaps with less carbon footprint than fine-tuned models or larger LLMs that require GPUs for training and/or inference. We do not foresee any harms due to the approaches described in this paper and it is only helpful for those working on text classification in the real-world to give a more realistic perspective about working with LLMs, which can potentially save time/money in a short/long run in terms of choosing the solution space to explore. We also share all the code and generated synthetic datasets as supplementary material to support reproducible research. 

\section*{Acknowledgments}
We thank Gabriel Bernier-Colborne for his feedback on an earlier version of this paper. This research was conducted at the National Research Council of Canada, thereby establishing a copyright belonging to the Crown in Right of Canada, that is, to the Government of Canada.

\bibliography{acl_latex}

\appendix

\section{Details about the Experimental Settings}
\label{sec:appendix}

\paragraph{Compute Infrastructure: }
All the zero-shot prompting, few-shot fine-tuning, and logistic classifier training, and synthetic data generation experiments were performed on an Apple M1 Pro laptop with 32GB memory. Open LLMs were downloaded and used locally using Ollama (\url{https://ollama.com/}), and the OpenAI model was called with their API. BERT-Finetuning for the datasets with smaller number of categories (\textsc{Sentiment} and \textsc{Taxi1500}) was done locally and for the other two datasets, it was done on a Google Colab T4 GPU. Instruction fine-tuning was performed on a V100 GPU. Transformers\footnote{\url{https://github.com/huggingface/transformers}} and Unsloth\footnote{\url{https://github.com/unslothai/unsloth}} were used for BERT-finetuning and instruction fine-tuning of Qwen2.5-7B respectively. All the implementation code is provided in the supplementary material. 

\paragraph{Zero-shot Prompting: }
Zero-shot prompting controlling for the output structure was performed using Instructor (\url{https://python.useinstructor.com/}), which utilizes Pydantic (\url{https://docs.pydantic.dev/}) for efficient data validation. The code snippet to prompt an LLM with instructor is shown in Figure~\ref{fig:prompt} below. The variable catnames contains the list of labels. More details on how structured output generation works can be seen in the code submitted as supplementary material and in the documentation for the library Instructor. 

\begin{figure}[htb!]
  \includegraphics[width=\columnwidth]{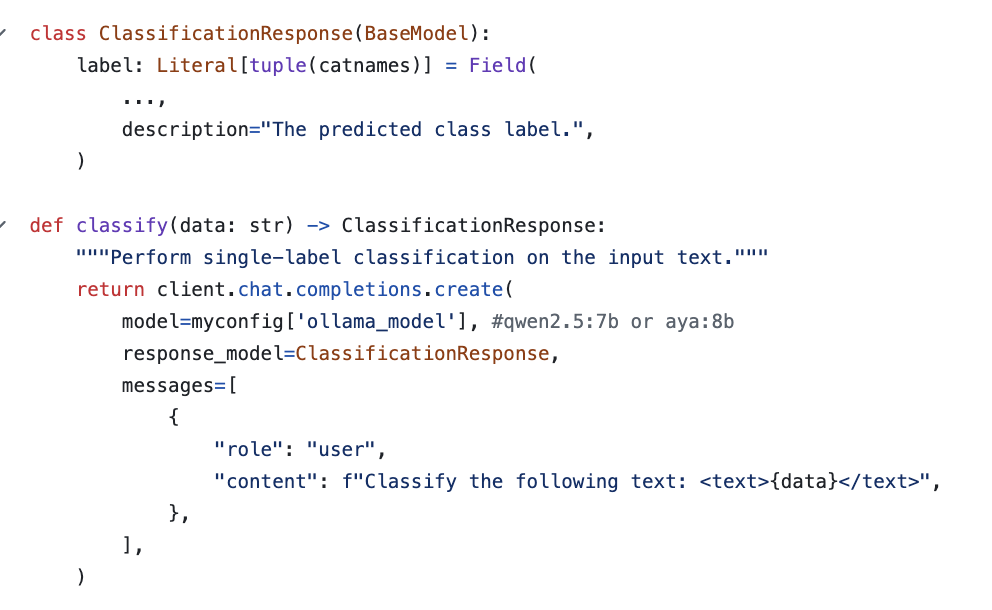}
  \caption{Prompt for Zero-shot Classification}
  \label{fig:prompt}
\end{figure}

\paragraph{Synthetic Data Generation: }
The prompt used for synthetic data generation is specified as follows, where \textit{text\_lang},  \textit{text\_genre}, \textit{task\_desc} and list\_of\_cats[i] come from config files and are task-dataset specific. For example, for generating sentiment classification data for Arabic, text\_lang = arabic, text\_genre = tweets, task\_desc = sentiment classification, list\_of\_cats = positive or negative or netural.    \\ \\
\begin{minipage}{0.45\textwidth}
\begin{verbatim}
prompt = f"""generate a {text_lang} language 
text that looks like {text_genre} 
that can be categorized as {list_of_cats[i]} 
in the context of {task_desc}. 
The generated text should be under 50 words, 
and ensure some diversity of vocabulary 
in the generated texts.
            """
\end{verbatim}
\end{minipage}

\paragraph{Instruction fine-tuning: }
The following instruction format was used for fine-tuning the Qwen2.5-7B model:
\begin{minipage}{0.45\textwidth}
\begin{verbatim}
''' <s>[INST] Consider the text: 
    "{input_text}" Please select the 
    most relevant category for the 
    given text from following OPTIONS:
    {all_categories}.        
    CHOICE: {response} </s>
'''
\end{verbatim}
\end{minipage}

\paragraph{Parameter settings: }
\label{app:settings}
We used evaluation loss as the metric for selecting the best model. Some training parameters are presented in Table \ref{tab:parameter-setting-for-FT}. Instruction tuning was done for 3 epochs each for \textsc{Sentiment}, \textsc{Intent} and \textsc{Scenario} datasets, and 10 epochs for \textsc{Taxi1500} dataset which has a smaller amount of training data compared to the rest and did not converge in 3 epochs. 

\begin{table}[htb!]
    \centering
    \small
    \begin{tabular}{|p{1cm}|p{0.5cm}|p{1cm}|p{1cm}|p{2cm}|}
    \hline
         Model &  epochs & learning rate & weight decay & GPU/CPU hours \\
         \hline
         FastFit & 10 & 5e-5 & 0.01 & 5-15 min for training; fast inference \\ \hline
         BERT & 5 & 5e-5 & 0.01 & $~$0.5 hr for training; fast  inference\\
         \hline
         Qwen2.5-7B & 3 & 1e-5  & 0.001 & $~$6hr for training; $~$3hr for inference\\
         \hline
    \end{tabular}
    \caption{Parameter setting for fine tuning}
    \label{tab:parameter-setting-for-FT}
\end{table}

More details can be seen in the code provided as supplementary material. 

\section{Additional Results}
\label{app:moreresults}

\begin{figure}[htb!]
  \centering
  \subfigure[\textsc{FastFit}]{\includegraphics[width=0.45\textwidth]{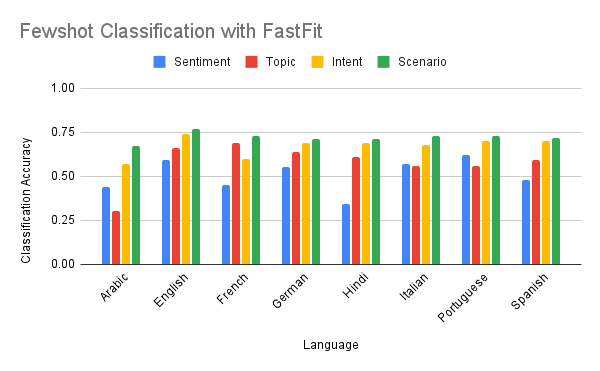}}
  \subfigure[\textsc{SetFit}]{\includegraphics[width=0.45\textwidth]{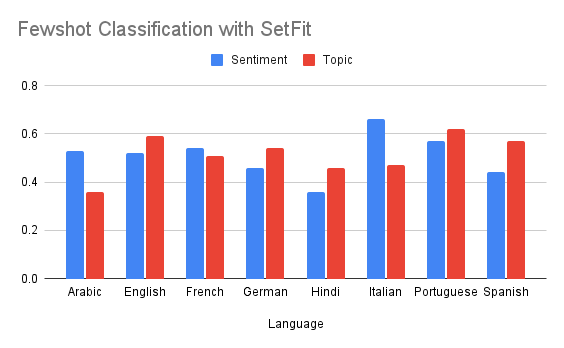}}
  \caption{Few-shot Fine-tuning Across Tasks and Languages}
  \label{fig:fewshotdetails}
\end{figure}




\begin{table*}[htb!]
\resizebox{\textwidth}{!}{
\begin{tabular}{|l|l|l|l|l|l|l|l|l|l|l|l|} \hline 
 & \multicolumn{4}{c|}{Zero-shot} & \multicolumn{1}{c|}{Few-shot} & \multicolumn{3}{c|}{Real Data}& \multicolumn{3}{c|}{Synthetic Data} \\ \hline
\textbf{Language} & GPT4 & Aya:8B & Qwen2.5:7B & Aya-Expanse:8B & FastFit & Logistic Reg. & BERT-FT & SFT & Logistic Reg. & BERT-FT & SFT\\ \hline 
Arabic & \textbf{0.68}& 0.59 & 0.58& \textbf{0.68}& 0.44  & 0.61  & 0.45  & 0.65& 0.51  & 0.5& 0.53\\
English& \textbf{0.71}& 0.65 & 0.68& \textbf{0.71}& 0.59  & 0.65  & 0.65  & 0.58& 0.59  & 0.52  & 0.51\\
French & 0.54 & 0.52 & 0.47& 0.58& 0.45  & 0.65  & \textbf{0.68} & 0.57& 0.61  & 0.49  & 0.45\\
German & 0.67 & 0.62 & 0.65& 0.7 & 0.55  & 0.7& 0.68  & \textbf{0.73}& 0.57  & 0.47  & 0.6 \\
Hindi  & \textbf{0.62}& 0.53 & 0.56& 0.61& 0.34  & 0.41  & 0.33  & 0.36& 0.47  & 0.38  & 0.46\\
Italian& \textbf{0.72}& 0.58 & 0.6& 0.71& 0.57  & 0.59  & 0.59  & 0.64& 0.61  & 0.49  & 0.59\\
Portuguese  & 0.69 & 0.6  & 0.61& 0.69& 0.62  & \textbf{0.7}& 0.54  & 0.68& 0.58  & 0.49  & 0.59\\
Spanish& 0.68 & 0.59 & 0.62& \textbf{0.69}& 0.48  & 0.65  & 0.56  & 0.61& 0.61  & 0.49  & 0.52 \\ \hline 
\end{tabular}}
\caption{All Results for \textsc{Twitter} sentiment classification task}
\label{tab:twitter-details}
\end{table*}

\begin{table*}[htb!]
\resizebox{\textwidth}{!}{
\begin{tabular}{|l|l|l|l|l|l|l|l|l|l|l|l|} \hline 
 & \multicolumn{4}{c|}{Zero-shot} & \multicolumn{1}{c|}{Few-shot} & \multicolumn{3}{c|}{Real Data}& \multicolumn{3}{c|}{Synthetic Data} \\ \hline
\textbf{Language} & GPT4 & Aya:8B & Qwen2.5:7B & Aya-Expanse:8B & FastFit & Logistic Reg. & BERT-FT & SFT & Logistic Reg. & BERT-FT & SFT\\ \hline 
Arabic& 0.58  & 0.22 & 0.23& 0.44& 0.3  & 0.36  & 0.5 & \textbf{0.69} & 0.32  & 0.48& 0.61 \\
English & 0.53  & 0.35 & 0.4 & 0.43& 0.66 & 0.74  & \textbf{0.77}& 0.69 & 0.54  & 0.51& 0.33 \\
French& 0.54  & 0.23 & 0.34& 0.41& 0.69 & 0.57  & \textbf{0.73}& 0.67 & 0.55  & 0.5 & 0.62 \\
German& 0.44  & 0.32 & 0.33& 0.41& 0.64 & 0.54  & \textbf{0.72}& \textbf{0.72} & 0.44  & 0.43& 0.57 \\
Hindi & 0.52  & 0.27 & 0.35& 0.44& 0.61 & 0.5& \textbf{0.64}& 0.62 & 0.46  & 0.41& 0.57 \\
Italian & 0.49  & 0.27 & 0.32& 0.38& 0.56 & 0.56  & 0.35& \textbf{0.6}& 0.48  & 0.49& 0.43 \\
Portuguese & 0.47  & 0.3& 0.41& 0.45& 0.56 & 0.55  & \textbf{0.69}& 0.56 & 0.58  & 0.48& 0.47 \\
Spanish & 0.56  & 0.23 & 0.4 & 0.52& 0.59 & 0.47  & \textbf{0.72}& 0.56 & 0.52  & 0.48& 0.37 \\ \hline
\end{tabular}}
\caption{All Results for \textsc{Taxi1500} Bible topic classification task}
\label{tab:taxi1500-details}
\end{table*}

\begin{table*}[htb!]
\resizebox{\textwidth}{!}{
\begin{tabular}{|l|l|l|l|l|l|l|l|l|l|l|l|} \hline 
 & \multicolumn{4}{c|}{Zero-shot} & \multicolumn{1}{c|}{Few-shot} & \multicolumn{3}{c|}{Real Data}& \multicolumn{3}{c|}{Synthetic Data} \\ \hline
\textbf{Language} & GPT4 & Aya:8B & Qwen2.5:7B & Aya-Expanse:8B & FastFit & Logistic Reg. & BERT-FT & SFT & Logistic Reg. & BERT-FT & SFT\\ \hline 
Arabic& 0.6& 0.31& 0.46& 0.49& 0.57& 0.74& 0.79& \textbf{0.81}& 0.44& 0.3& 0.51\\
English& 0.72& 0.44& 0.6& 0.6& 0.74& 0.87& 0.88& \textbf{0.89}& 0.58& 0.43& 0.62\\
French& 0.67& 0.4& 0.54& 0.56& 0.6& 0.83& 0.86& \textbf{0.87}& 0.58& 0.46& 0.56\\
German& 0.66& 0.37& 0.54& 0.56& 0.69& 0.8& 0.84& \textbf{0.86}& 0.51& 0.43& 0.58\\
\textbf{Hindi} & 0.66& 0.39& 0.5& 0.52& 0.69& 0.83& 0.82& \textbf{0.86}& 0.57& 0.45& 0.59\\
Italian& 0.68& 0.41& 0.55& 0.56& 0.68& 0.83& 0.85& \textbf{0.86}& 0.57& 0.43& 0.57\\
Portuguese     & 0.66& 0.38& 0.54& 0.55& 0.7& 0.83& 0.86& \textbf{0.88}& 0.56& 0.44& 0.59\\
Spanish& 0.66& 0.37& 0.54& 0.54& 0.7& 0.84 &\textbf{0.86} &0.85&0.54 &0.44&0.61   \\ \hline 
\end{tabular}}
\caption{All Results for \textsc{Intent} classification task}
\label{tab:intent-details}
\end{table*}

\begin{table*}[htb!]
\resizebox{\textwidth}{!}{
\begin{tabular}{|l|l|l|l|l|l|l|l|l|l|l|l|} \hline 
 & \multicolumn{4}{c|}{Zero-shot} & \multicolumn{1}{c|}{Few-shot} & \multicolumn{3}{c|}{Real Data}& \multicolumn{3}{c|}{Synthetic Data} \\ \hline
\textbf{Language} & GPT4 & Aya:8B & Qwen2.5:7B & Aya-Expanse:8B & FastFit & Logistic Reg. & BERT-FT & SFT & Logistic Reg. & BERT-FT & SFT\\ \hline 
Arabic & 0.59& 0.43& 0.52& 0.45& 0.672& 0.58& \textbf{0.86}& \textbf{0.86}& 0.58& 0.54& 0.696 \\
English & 0.67& 0.5 & 0.66& 0.52& 0.77& 0.9 & 0.9 & \textbf{0.94}& 0.74& 0.54& 0.77 \\
French & 0.64& 0.47& 0.59& 0.49& 0.73& 0.81& \textbf{0.9} & \textbf{0.9} & 0.7 & 0.59& 0.72  \\
German & 0.64& 0.46& 0.6 & 0.5 & 0.71& 0.79& \textbf{0.9} & \textbf{0.91}& 0.68& 0.59& 0.82\\
Hindi & 0.62& 0.45& 0.54& 0.48& 0.71& 0.55& \textbf{0.87}& \textbf{0.87}& 0.7 & 0.57& 0.79\\
Italian & 0.63& 0.46& 0.59& 0.49& 0.73& 0.8 & \textbf{0.89}& \textbf{0.89}& 0.68& 0.61& 0.75\\
Portuguese & 0.63& 0.46& 0.61& 0.5 & 0.73& 0.81& 0.9 & \textbf{0.93}& 0.67& 0.57& 0.76\\
Spanish & 0.63& 0.46& 0.6 & 0.48& 0.72& 0.79& 0.89& \textbf{0.9} & 0.68& 0.56& 0.74\\ \hline 
\end{tabular}}
\caption{All Results for \textsc{Scenario} classification task}
\label{tab:scenario-details}
\end{table*}

\end{document}